\documentclass[conference]{IEEEtran}

\usepackage{cite}
\usepackage{amsmath, amssymb, amsfonts}
\usepackage{graphicx}
\usepackage{textcomp}
\usepackage{xcolor}
\def\BibTeX{{\rm B\kern-.05em{\sc i\kern-.025em b}\kern-.08em
    T\kern-.1667em\lower.7ex\hbox{E}\kern-.125emX}}
\usepackage[linesnumbered,ruled,vlined]{algorithm2e}
\usepackage[small]{caption}
\usepackage{subfig}
\usepackage{color,soul}
\usepackage{array}
\usepackage{booktabs}
\usepackage{footnote}
\makesavenoteenv{tabular}
\makesavenoteenv{table}
\usepackage[linesnumbered,ruled,vlined]{algorithm2e}
\usepackage[noend]{algpseudocode}
\usepackage{float}
\usepackage[utf8x]{inputenc}    % For the Scandinavian characters
\usepackage{textcomp}   % Degree Celsius 
\usepackage{mathtools}

\usepackage{multicol}
\usepackage{graphicx}

\renewcommand\baselinestretch{0.92}
\begin{document}
\bstctlcite{IEEEexample:BSTcontrol}
\title{Efficient Training of Deep Convolutional Neural Networks by Augmentation in Embedding Space}

\author{
\IEEEauthorblockN{Mohammad Saeed Abrishami, Amir Erfan Eshratifar, Yanzhi Wang,Shahin Nazarian, David Eigen, Massoud Pedram\\}
\IEEEauthorblockA{Ming Hsieh Department of Electrical and Computer Engineering\\
Viterbi School of Engineering, University of Southern California \\ 
Los Angeles, CA 90089 }}

\author{    Mohammad Saeed Abrishami\textsuperscript{1}, \and
            Amir Erfan Eshratifar\textsuperscript{1}, \and
            David Eigen\textsuperscript{3}, \and
            Yanzhi Wang\textsuperscript{2}, \and
            Shahin Nazarian\textsuperscript{1}, \and
            Massoud Pedram\textsuperscript{1}\\
\textsuperscript{1}{Department of Electrical and Computer Engineering at University of Southern California}\\
\textsuperscript{2}{Department of Electrical and Computer Engineering at Northeastern University}\\
\textsuperscript{3}{Clarifai Inc, San Francisco}\\
\{abri442, eshratif\}@usc.edu,
deigen@clarifai.com,
yanz.wang@northeastern.edu,
\{s.nazarian, pedram\}@usc.edu}

\maketitle
\thispagestyle{plain}
\pagestyle{plain}
\begin{abstract}
% 1. State the problem
% 2. Say why it’s an interesting problem
% 3. Say what your solution achieves
% 4. Say what follows from your solution 
Recent advances in the field of artificial intelligence have been made possible by deep neural networks. In applications where data are scarce, transfer learning and data augmentation techniques are commonly used to improve the generalization of deep learning models. However, fine-tuning a transfer model with data augmentation in the raw input space has a high computational cost to run the full network for every augmented input. This is particularly critical when large models are implemented on embedded devices with limited computational and energy resources. In this work, we propose a method that replaces the augmentation in the raw input space with an approximate one that acts purely in the embedding space. Our experimental results show that the proposed method drastically reduces the computation, while the accuracy of models is negligibly compromised.
\end{abstract}

% The recent drastic improvements in the field of artificial intelligence have been made possible by deep neural networks. Promising results expanded the deployment of these new learning architectures to different applications, such as surveillance and autonomous driving. In data-scarce applications, transfer learning and data augmentation techniques are commonly used to improve the generalization of the models. However, the status-quo in fine-tuning a transferred model with augmentation comes with a high computation cost of feed-forwarding the complete network when it is applied in the raw pixel space. However, fine-tuning a transferred model with augmentation in the raw input space comes with the high computational cost of running the full-network for every augmented input. This is more critical when large models are implemented on embedded devices with limited computational and energy resources. In this work, we propose a method that replaces the augmentation in the raw input space with the one in the embedding space which reduces the computations during transfer learning. Our experimental results show that the proposed method drastically reduces the computation while the accuracy of models is negligibly compromised.

\IEEEpeerreviewmaketitle

\section{Introduction}\label{sec:intro}
% Describe the problem
% State your contributions
% Use an example to describe your problem
% Your introduction makes claims. 
% The body of the paper provides evidence to support each claim
% Check each claim in the introduction, identify the evidence, and forward reference it from the claim
% "Evidence" can be: analysis and comparison, theorems, measurements, case studies

% Paragraph on the application of DNNs
Deep learning is one of the main elements of the recent advances in the field of artificial intelligence. 
Some of the major factors accelerating the progress of such models can be listed as:
significant increase in the amount of available training data, the evolution of computational power of electronic devices, and introducing new learning algorithms and open-source tools~\cite{deep-learning-nature}.
The superiority of \textit{deep neural networks} (DNNs) to other methods was initially presented by setting records in well-known challenging artificial intelligence tasks, such as image classification~\cite{alexnet}, speech recognition\cite{deep-speech}, etc. 
Especially, with the increase of accessibility to mobile devices, many of these tasks are running on embedded systems. 
% TODO: do we need NLP?

% Paragraph on computation 
Improvement of the model's accuracy was considered as a top priority objective in the early-stage deep learning research and thus resulted in the appearance of computational-hungry models. 
Even with the drastic computation capability improvement of graphical processing units (GPUs), which are known as the common practice platforms for training DNNs, training advanced DNN models may take several hours to multiple days~\cite{Ng-ML-GPU}. 
However, there are major shortcomings when DNN models are deployed on embedded devices:
1) the computational capabilities of such devices are very limited, and 
2) embedded devices are mostly battery-based and have energy consumption constraints even for simple tasks~\cite{dnn-practical}. 
Therefore, the training process of DNNs is typically offloaded to the cloud as it requires a large amount of computation on large datasets. 
Once the model is trained, it will be used for inference on new unseen inputs. The inference process can be hosted privately on the local devices or as a public service on the cloud.

%% The computational power of embedded devices has been enhanced dramatically in the last few years. 
% Moreover, considering the broad applications of AI, they are recently being equipped with specific hardware such as high-performance on-chip GPUs and \textit{system-on-chips} (SoCs) aimed to accelerate the performance of machine learning tasks. 
% However, their computational power is generally not comparable with cloud servers. 
%% It should be mentioned energy consumption is always an important factor in using mobile devices as they are dependent on constrained energy batteries~\cite{ai-edge}. 
%% While these suggest offloading the training and inference queries to the cloud as an optimized decision, there are a few shortcomings. 

%% Reliability on the network connection and high latency for applications such as autonomous vehicles and drone navigation are other shortcomings of offloading to the cloud. 
%% For many applications such as surveillance face recognition and traffic control, it is, in fact, necessary to locate the AI-engine adjacent to the image sensor rather than in the cloud due to privacy and security concerns.
%% Therefore, low-cost on-device computation is challenging but yet critical for many applications. 

% Why device
The communication cost of cloud-based inference can be also larger than the computation cost of running a small model locally. Collaborative approaches between the cloud, edge, and the mobile devices are proposed to co-optimize the communication and computation costs simultaneously~\cite{JointDNN, edge-glsvlsi}. 
While cloud-based inference is easy to deploy and scale up, it compromises the data privacy and needs a reliable network connection. 
In some mission-critical applications running on embedded devices, such as drone navigation, it is required for the model to continuously improve or adapt to new unseen tasks. 
This emphasizes the requirement of reliable and privacy-preserving setups for training, which cannot be satisfied with cloud-based approaches. 
Therefore, for some applications running on resource-constrained devices, local training and inference are needed.

% Transfer learning and Data augmentation
The generalization performance of DNNs is challenging because of the possible distribution misalignment between the training and test sets. 
Overfitting, i.e., learning too much from the training set, prevents the DNN model from performing well in unseen real environments despite the high accuracy on the training set. 
This is why well-known problems such as image classification are trained with millions of trained samples. 
However, for many applications, large labeled datasets are either unavailable or very expensive to annotate for training a model with a specific task. 
To mitigate the limited number of samples in practical settings, transfer learning~\cite{transfer-learning, Oquab2014LearningAT} and data augmentation~\cite{auto-augment-google} are effective methods to improve the generalization of the learning model. 
Transfer learning is introduced to create pre-trained models on datasets that are considered similar to the dataset of the final task.
As the initial model is trained only once with a large amount of data, the training can be done on cloud in which the computational cost is not a major concern. 
Later, the pre-trained model with optimized parameters is fine-tuned to fit the new problem. 
Data augmentation is another strategy that helps to increase the diversity of the training data without collecting new data. 
For instance, in computer vision applications, augmentation techniques such as mirroring the input image are commonly used to improve generalization. The main disadvantage of augmentation is that it enforces a linear increase
in the number of feed-forward calls with respect to the number of augmentations used for fine-tuning. 

% Paragraph finale 
In this paper, we present a novel idea that drastically reduces the computation work required for fine-tuning DNNs by augmenting the input in the embedding space instead of the raw input space. The paper makes the following contributions:
\begin{itemize}
    \item Introducing a novel method for augmentations in the embedding space instead of raw input space. 
    \item Analysis of the impact of our method on the computation and accuracy of transferred models with different network architectures. 
\end{itemize}

% On the other hand, the local inference enables the mobile application to function without network access but is limited to small models due to the lack of enough computing resources. 
% Also, for mission-critical applications, it is desirable for the edge devices to be able to fine-tune its model on its own. As a result, reducing the computation costs of training on embedding devices is also important.

% Model compression techniques have been proposed to reduce the computational demand often by trading the prediction accuracy. 
% These techniques include quantization \cite{ Han:2015:LBW:2969239.2969366, Rastegari2016XNORNetIC}, 
%pruning \cite{ Han:2016:EEI:3007787.3001163}, 
%optimized convolution operations \cite{ squeezenet,  Howard2017MobileNetsEC,Georgiev2017LowresourceMA}, 
%and knowledge distillation for training small models using the knowledge of a teacher model \cite{Hinton2015DistillingTK}. 
% Hardware-aware neural architecture search is also a recent interesting and promising research area \cite{Tan2018MnasNetPN}. 
% \textcolor{red}{ Other researchers} explore the space of hybrid approaches that splits the computations between the edge devices and cloud servers. For instance, JointDNN\cite{JointDNN} decides to offload some, or all layers computations in a DNN from an edge device to the cloud servers for reduced latency and energy consumption for both training and inference phases.

\section{Background and Related Works} \label{sec:background}
% \begin{itemize}
    % \item Huge computation of DNNs
    % \item Platform of computation of DNNs for training and inference, cloud, embedding
    % \item What is transfer learning
    % \item Introduce the terminology in the paper
    % \item Augmentation, its benefits, then add the computation required for augmentation
% \end{itemize}

\begin{figure}
    \centering
    \includegraphics[width=\columnwidth]{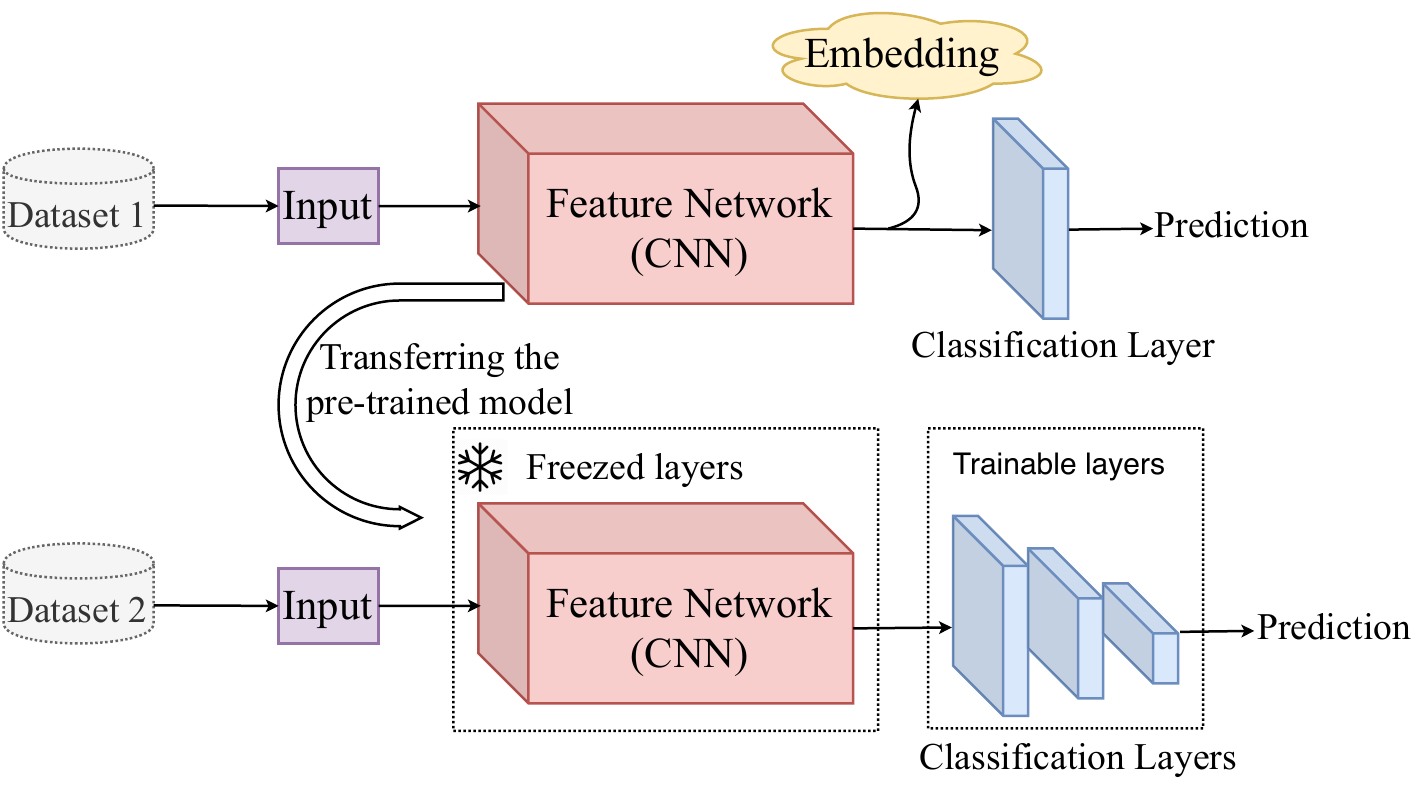}
    \caption{Transfer learning in deep neural networks. The feature network is pre-trained on one dataset then its parameters are copied and freezed (non-trainable) for learning on the second dataset. Only the added downstream classification layers are trainable for the second dataset.} \label{fig:transfer_learning}
\end{figure}

% Concept of transfer learning
Transfer learning~\cite{transfer-learning} is known as the status-quo approach in data-scarce scenarios. 
In this method, a base-model is first pre-trained for a data-rich task, referred to as the base-task. 
Throughout extensive training, the base-model learns how to extract high-level features from the base-dataset. 
These high-level features are referred to as \textit{embedding} and are typically generated at the final layers of the DNN. 
The advantage of transferring knowledge to a downstream model is not only limited to higher accuracy, but also fast convergence and lower training computation. 

% Details of transfer learning
The main idea of transfer learning for a data-scarce task is to transfer the mature feature extraction part of a pre-trained base-model to a target-network and fine-tune it on the target-dataset to fit the target-task. 
It is common practice to only update the parameters of the downstream layers after the embedding during fine-tuning and keep the other parameters fixed (frozen). 
The transfer learning procedure in DNNs is illustrated in Fig.~\ref{fig:transfer_learning}. 
In cases with very limited target-dataset samples, the problem is referred to as \textit{few-shot learning}~\cite{one-shot-learning, meta-learning-aaai}. 
The quality of such approaches is dependent on the extent of similarity between the distribution of the base and target tasks~\cite{tf-similarity, transfer-learning}.

% Augmentation
Lack of sufficient labeled data may cause overfitting due to sampling bias in DNNs. 
Data augmentation is known as a powerful method to reach higher generalization and prevent overfitting by simply inflating the training data size. New samples are reproduced from a single sample while the label is not changed or is known without any further annotation~\cite{data-augmentation}. 
In particular, for the case of image classification tasks~\cite{image-augmentation-effectiveness, population-augmentation, auto-augment-google}, every training image sample can be modified by applying transformations such as horizontal and vertical flip, image rotation, random cropping, perturbations to brightness, contrast, color, etc.

% Augmentation literature review
% invariance to translation
One of the useful characteristics of \textit{convolutional neural networks} (CNNs) is their translation equivariance. 
In other words, translating the input image is the same as translating the feature maps, due to the symmetry preserving characteristics of each layer. 
The equivariance relationship of other augmentations such as flips, scaling and rotation are further studied in ~\cite{equivariance-equivalence-cvpr2015} by finding a relationship between representations of the original and transformed images. 
Furthermore, the operations in CNNs are extended in~\cite{group-equivariant-mlr} to be formally equivariant to reflections. 
The impact of equivariance relationship has been extended to time-series data such as videos~\cite{ego-motion}. 
The main difference between our proposed method and the prior works is that instead of enforcing the model to be equivariant to augmentations, we learn the transformations that map the embedding of the original input to the augmented ones.

%tasks~\cite{Devlin2019BERTPO}. % ~\cite{Kiela2014LearningIE} 

\section{Methodology}\label{sec:methodology}
% TODO: differentiate with training examples and fine-tuning examples. 
In the following section, we introduce our proposed method, i.e. replacing augmentation in the pixel space with one in the embedding space. 
Moreover, we elaborate upon on how this new idea saves computation when transferring models. 

\subsection{Embedding}
\textit{Deep convolutional neural networks} (DCNNs) are comprised of several convolutional blocks including convolutional filters, pooling, activation functions, etc., followed by one or a few \textit{fully connected} (FC) layers. 
Although there are still unanswered questions on the profound results of DCNNs, the belief is that multiple convolutional layers learn the intermediate and high-level features in different levels of abstraction between the input image and the output~\cite{deep-shallow-2017, DCNN-feature-2018}.

% TODO: 1-D should be defined as 1-dimensional
In most computer vision applications, the embedding is a  1-dimensional vector with continuous values, in floating point.
In practice, the output of the last convolutional layer and before the first FC layer is commonly considered as the embedding. 
While there is no certain understanding of what does a single feature represents in an embedding vector, they can meaningfully represent the semantic features of an image in a transformed space. 
% Maybe a reference is required here.
In classification networks, the FC layers at the end of the network are responsible for mapping this embedding into different classes. 
Throughout this paper, we present the 
\textit{feature generator sub-network} with $\Phi$, 
the \textit{classifier sub-network} with $\Psi$, 
and the embedding vector with $z$. 
% The neural network model is in fact a nonlinear function applied on input images. 
The notations used in this paper are summarized in Table~\ref{tab:notation}. 
The relationship between input, embedding, output, and the two sub-networks are given below: 

% TODO: make two equations here
\begin{align}\label{eq:PhiShi}
    z_i &= \Phi(x_i) \\
    \hat{y}_i &=  \Psi(z_i) = \Psi (\Phi(x_i))
\end{align}

\begin{table}[]
\begin{tabular}{|c|p{195pt}|}
\hline
Notation & Description \\ \hline
$x_i \in X$ & Original image dataset and one image sample data, without any augmentation \\ \hline
$y_i$ & Label of image $x_i$ \\ \hline
$\hat{y}_i$ & Predicted label of image $x_i$ \\ \hline
$g_j \in G$ & Augmentation function set \\ \hline
$x_{ij}$ & Image $x_i$ under augmentation $g_j$ in pixel space \\ \hline
$\Phi$ & Feature generator sub-network, usually convolutional layers \\ \hline
$\Psi$ & Classifier sub-network, usually FC layers \\ \hline
$z_i$ & Embedding of image $x_i$, usually a 1-D vector \\ \hline
$z_{ij}$ & Embedding of augmented image $x_{ij}$\\ \hline
$\Omega_j$ & Augmentation transformer for $g_{j}$\\ \hline
\end{tabular}
\caption{Description of the notations used in this manuscript.} 
\label{tab:notation}
\end{table}

\begin{figure*}[t]
\centering
\includegraphics[width=\linewidth]{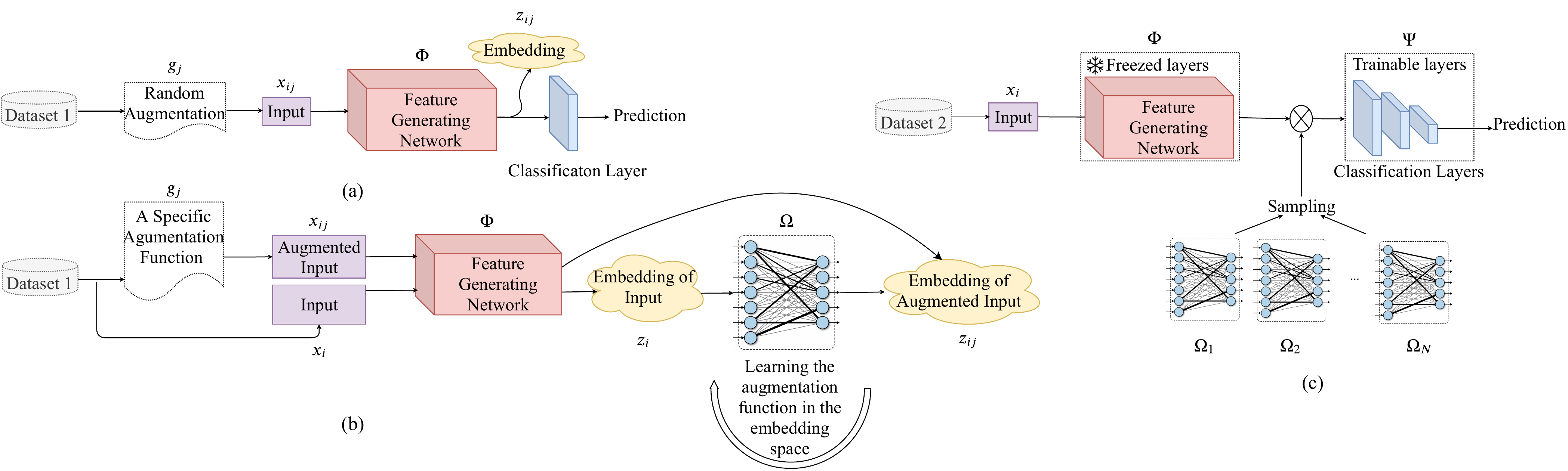}
\caption{Augmentation in the embedding space. Our procedure follows three steps: (a) a large model (feature network) is trained on the source dataset with all the augmentation functions. (b) pairs of the embedding of input and its augmented one are extracted from the feature network and used as the training set for learning the augmentation functions in the embedding space ($\Omega$). (c) The learned augmentation functions in the embedding space are used to accomplish the transfer learning for the target dataset.}
\label{fig:augmentation_embedding}
\end{figure*}

\subsection{Augmentation in the pixel space}
We present a scenario to illustrate the functionality of the proposed method. 
A DCNN model is implemented on a platform with limited computational and energy resources, such as a smartphone. 
The target-model is initially transferred from a base model, which is pre-trained on a separate platform, such as a cloud provider. 
This transferred model is designated to do a specific task, thus fine-tuning with new sample images is required. 
Using augmented images is crucial to improving the accuracy, however, it comes with the cost of higher computation. 
Specifically, generating $\mathrm{N}$ augmented images from the original one for fine-tuning will increase the total computation of $\mathrm{N}\times$, which includes feed-forward, and optionally, the more complex back-propagation phases for the augmented samples. 
We refer to this method as \textit{"augmentation in the pixel space"} as the new images are augmented by techniques such as horizontal flip (mirroring), vertical flip, rotation, crop, etc. applied on the original image as illustrated in Fig.~\ref{fig:augmentation_embedding}-a. 
The new relationships can be written as follows: 

% TODO: make two equations here
\begin{align}\label{eq:PhiPsi-aug}
    x_{ij} &= g_j(x_i) \\ 
    z_{ij} &= \Phi(x_{ij}) \\
    \hat{y}_{ij} &= \Psi(z_{ij})= \Psi (\Phi(x_{ij})) = \Psi (\Phi(g_j(x_i)))
\end{align}

\subsection{Augmentation in the embedding space}
The main part of our proposed method is that embedding of an augmented image ($z_{ij}$) can be generated directly by using the embedding of the original image ($z_i$).
In other words, $z_{i}$ cab be approximately transformed into $z_{ij}$ by a simple nonlinear function, referred to as \textit{augmentation transformers} and represented with $\Omega$.
Eq.~\ref{eq:omega} shows the functionality of $\Omega_j$ for the specific augmentation $g_j$. 

\begin{align}\label{eq:omega}
    \Omega_{j} (z_i) = \Omega_{j} (\Phi(x_i)) &\approx z_{ij} = \Phi(g_j(x_i))
\end{align}

A neural network with a few FC can be used to implement $\Omega$. 
Consequently, the parameters of $\Omega_j$ can be optimized by training on the same dataset and computation platform as the base model.
This process is done after the base model is completely trained and thus parameters of $\Phi$ are kept fixed for later transfer learning on target devices. 
The input and output of the training process of $\Omega_j$ are embeddings of original images ($z_i$s) and embeddings of augmented images in the pixel space ($z_{ij}$). 
Input image $x_i$ from base dataset is passed through $\Phi$ sub-network and its embedding ($z_i=\Phi(x_i)$) is recorded. 
In addition, an augmentation in the pixel space ($g_j$) is applied on the input image ($x_{ij}=g_j(x_i)$), passed through $\Phi$, and its embedding $z_{ij}=\Phi(x_{ij})$ is then collected. 
This process is repeated separately for different augmentations. 
The objective is to find the optimized parameters of each $\Omega_j$ neural network to improve the approximation in Eq.~\ref{eq:omega}. 
Mean squared error (MSE) is used as a simple measurement for similarity loss and the objective function can be formulated as Eq.~\ref{eq:mse}. The parameters can be optimized by applying back-propagation of this loss using any gradient descent based method. 

% TODO: we need to show that \Omega is a function of theta
\begin{equation}
\begin{aligned}\label{eq:mse}
J &= \underset{\theta}{\operatorname{argmin}} \frac{1}{n} \sum_{i=1}^{n} ||z_{ij} - \Omega_j (z_i) ||^2  \\ 
&= \underset{\theta}{\operatorname{argmin}} \frac{1}{n} \sum_{i=1}^{n} ||\Phi(g_j(x_i)) - \Omega_j(\Phi(x_i))||^2 
\end{aligned}
\end{equation}

The flow of training the augmentation transformations is illustrated in Fig.~\ref{fig:augmentation_embedding}-b. 
% This part can be removed
It should be emphasized that different augmentation transformations are implemented and trained with separate neural network models, even though they can have the same architectures. 
% TODO: talk about we did not use other datasets but used the same dataset as base-model.
This step can be done on the cloud and can be passed alongside the base model to be used for later transfer learning purposes. %TODO: add to this. 

\subsection{Computation analysis}
%TODO: show cost (C_\Phi, C_\Psi, and C_\Omega in the figure
% We need to explain the training of \Psi_{tl} as well and refer to fig2-c.
As shown in Fig.~\ref{fig:augmentation_embedding}, the required total computation of training when an image augmentation is applied in the pixel space can be listed as the computation of embedding ($C_{\Phi}$), embedding to output ($C_{\Psi}$), and finally back-propagation on the layers which are not frozen, which are only the FC layers ($C_{\Psi-BP}$). 
On the other hand, if augmentation is done in the embedding space, only the embedding of the original image are computed once for all different augmentations and then transformed using $\Omega$s for fine-tuning the model. 
The total computation in this scenario can be listed as transforming the embedding of the original image to the augmented one ($C_\Omega$), and similar to the first case $C_{\Psi}$ and $C_{\Psi-BP}$. 
Augmentation transformers ($\Omega$s) are simple nonlinear functions implemented with FC layers. 
The number of parameters in these layers and the required computation is expected to be much smaller than the ones for $\Phi$, therefore, $C_\Omega$ is expected to be relatively much lower than $C_{\Phi}$. 
The relative total computation saving achieved by applying augmentation in the embedding space ($C_{E}$) instead of pixel space ($C_{P}$) when $N$ different augmentations are applied is formulated in Eq.~\ref{eq:comp}:

\begin{align}\label{eq:comp}
    \frac{C_P}{C_E} = 
    \frac{N \times (C_\Phi + C_\Psi + C_{\Psi-BP})}{C_\Phi + N \times (C_\Omega + C_\Psi + C_{\Psi-BP})}
\end{align}

In this equation, we ignored the cost of augmentation in the pixel space. 
\section{Experiments and Simulation Results} \label{sec:experiments}

% \subsection{Experimental Setup}
%\textbf{Hardware.} We evaluate our approach on the NVIDIA Jetson TX2 embedded deep learning platform as our mobile device. The system has a 64 bit dual-core Denver2 and a 64 bit quad-core ARM Cortex A57 running at 2.0 GHz, and a 256-core NVIDIA Pascal GPU running at 1.3 GHz. The board has 8 GB of LPDDR4 RAM and 96 GB of storage (32 GB eMMC plus 64 GB SD card). \\

% TODO: Erfan please put them in the right place
The implementations are done with PyTorch\cite{pytorch}, cuDNN (v7.0) and CUDA (v10.1). We study our proposed method on image classification task and top-1 accuracy is reported.

\subsection{DCNN architectures and image datasets}
For better evaluation of our proposed method, we did our experiments with different types of state-of-the-art DCNN architectures. 
VGG-16~\cite{vgg} is an architecture with 16 layers including 13 convolutional layers proceeded by 3 FC layers. 
% MerC Erfan! mokhlesim
% RESNET: 
The family of ResNet~\cite{Resnet} architectures use identity shortcut connections that skip one or more layers. 
The main advantage of using residual blocks is overcoming the vanishing gradient problem in deep networks. 
The implementation of this architecture can have different deterministic depths. 
Considering our scenario on the implementation of our network on embedded devices, we chose ResNet-18 with 17 convolutional layers and a single FC layer at the end. 
%Salient part in different images can have large size variation. % TODO: grammar check
%A large convolutional kernel can extract information that is distributed more globally, on the other hand, a smaller one is preferred for information that is distributed more locally. 
%Therefore, this variation can potentially influence the functionality of convolutional kernels.
Inception-V3~\cite{inception-v1, inception-v23} is designed to have filters with multiple sizes to extract features even when the size of the salient part of the image is varying. 
We chose Inception-v3~\cite{inception-v23} in our implementations. 
% TODO: what about conv and fc number? 
The size of the embeddings is 512 for the VGG-16, ResNet-18 and 1028 for Inception-V3. 
CIFAR100 and CIFAR10~\cite{cifar} are chosen for the base and target datasets respectively. 
%All three network architectures are designed to perform classification tasks on the ImageNet dataset~\cite{imagenet}. 
%While we kept the general form of the networks as they were initially designed.
%However, we made a few minor modifications, such as changing the pooling sizes, to accommodate the difference between the sizes of the ImageNet images ($256\times 256$) and CIFAR images ($32 \times 32$).

We focus on horizontal and vertical flip augmentations as two of the widely used functions in computer vision applications. We have three different setups for the training of the base networks: 1) no augmentation 2) only horizontal flip, and 3) both horizontal and vertical flips to be applied in the pixel space. 
The results summarized in Table~\ref{tab:results} show the importance of augmentation as a horizontal flip improves the evaluation results by about 10\% on average. % TODO: check
As mentioned earlier, the augmentation is very dependent on the dataset, and as the results suggest, the vertical flip did not help the training but reduced the accuracy by about $2\%$. 
% vgg: 7.04%
% resnet: 12.59%
% inception: 10.33%
After this phase, the parameters of $\Phi$ are fixed to be used as our base network for training $\Omega$s and transfer learning. 

\subsection{Training augmentation transformers ($\Omega$)}
The embedding transformation functions are implemented as two FC layers with ReLU activation serving as the nonlinearity of the hidden layers. 
The input and output sizes of $Mg$-s are the same as the base network's embedding, and the size of the hidden layer is considered twice the input size. 
It should be mentioned a few other architectures such as deeper or wider ones were used, however, the results did not change significantly. 

\begin{figure}
    \centering
    \includegraphics[width=\columnwidth]{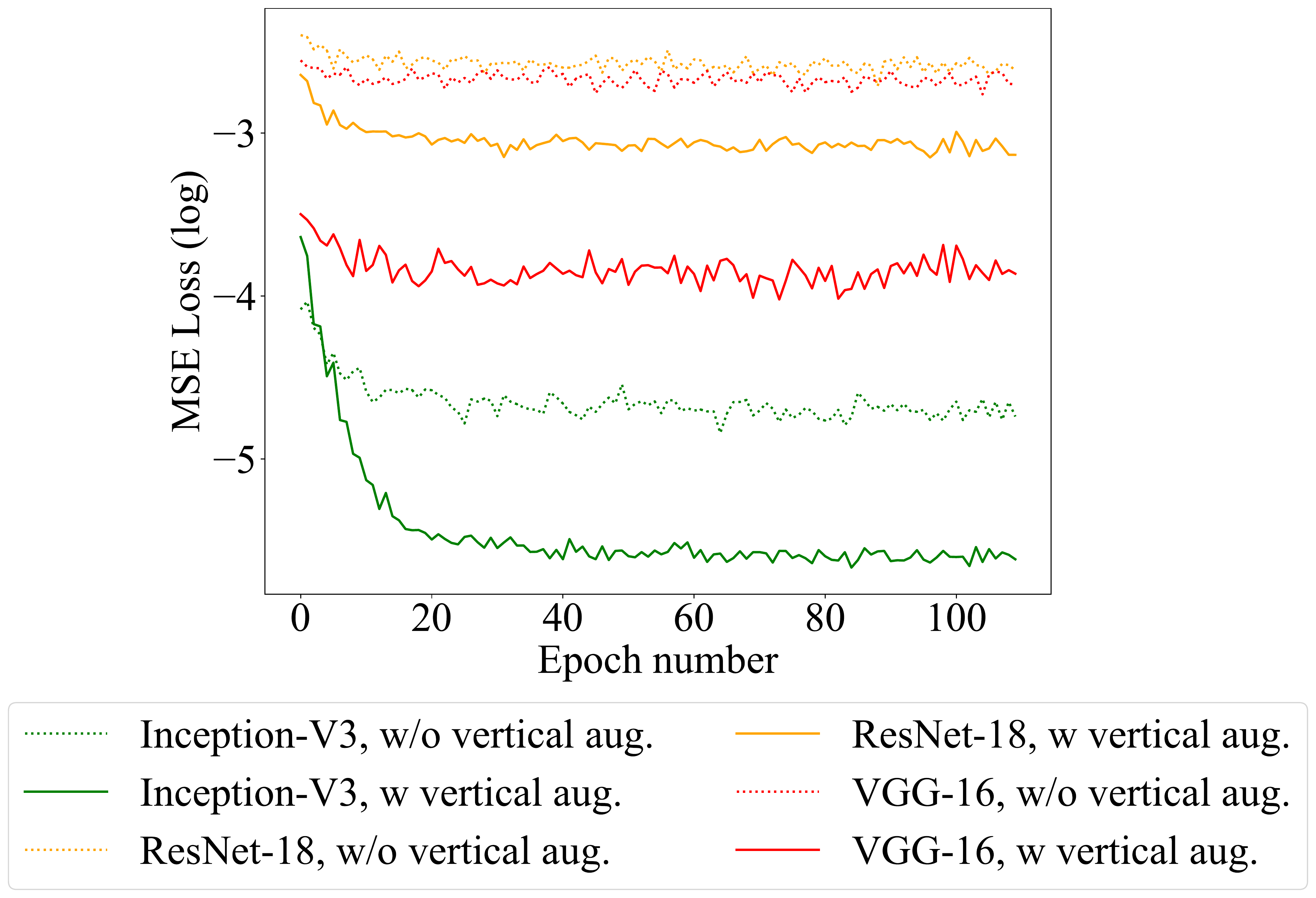}
    \caption{The MSE loss value of $\Omega$ which is trained on vertical augmentation when the base models is trained with and without the vertical augmentation. As we see, when the base model is not trained with the target augmentation, the loss value is higher.}
\label{fig:mg}
\end{figure}

The training data in this step is the same as the base-data, i.e. CIFAR100. To train each $\Omega_j$ for augmentation $g_j$, the training image $x_i$ and its augmented one $g_j(x_i)$ are fed through $\Phi$ to generate $z_i$ and $z_{ij}$, to be used as input and output respectively. 
As $\Omega$s are simple shallow networks, \textit{stochastic gradient descent} (SGD) optimization was used for back-propagation. 
As results demonstrated in Fig.~\ref{fig:mg} suggest, the evaluation loss of vertical augmentation transformer $\Omega_{ver}$ during training is much lower when the base network is trained with vertical augmentation in the pixel space. 
This emphasizes the hypothesis that the base network must be exposed to that specific augmentation so it can learn to generalize well. 

\subsection{Transfer learning}
\begin{figure*}
    \includegraphics[width=\linewidth]{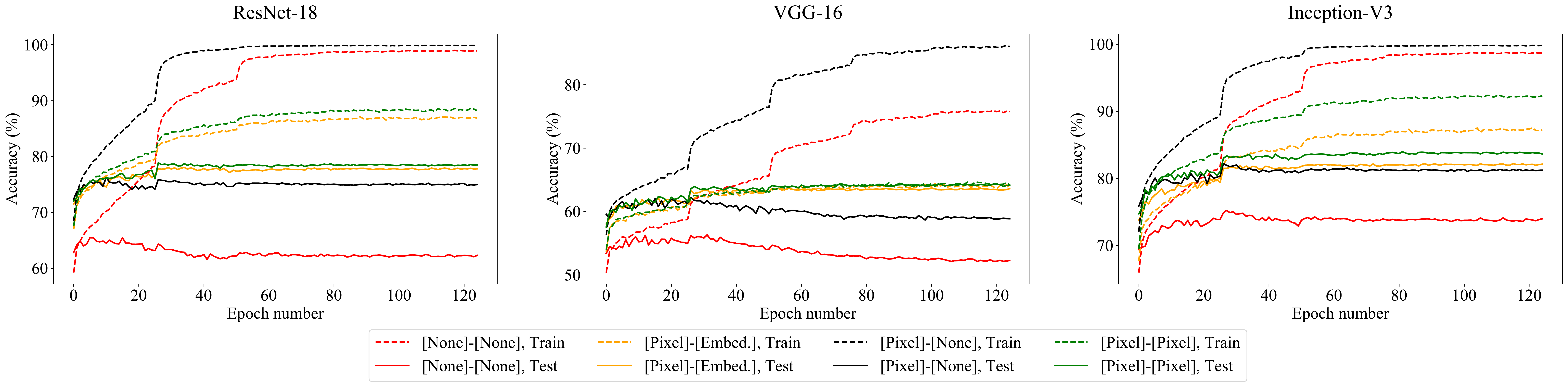}
\caption{The transfer learning accuracy results for three different deep models: (left) ResNet-18, (middle) VGG-16, (right) Inception-V3. Each figure shows four different scenarios: 1- The base and target models are both trained without any augmentation (red). 2- The base model is trained with pixel space augmentation and the target model is trained with embedding augmentation which is the proposed approach (yellow). 3- The base model is trained with pixel space augmentation and the target model is trained without any augmentation (black). 4- The base and target models are both trained with pixel space augmentation (green). }
\label{fig:transfer_learning_curves}
\end{figure*}

For transfer learning on CIFAR10, $\Psi$ is detached from the network and is replaced with 3 consecutive FC layers (1024, 128, and 10 neurons) serving as a classifier. 
Parameters of $\Phi$ are fixed and only the $\Psi$ is updated with back-propagation during fine-tuning. 
We have done four different experiments in this part. 
For the first case, the input images were not augmented either during the training of the baseline or fine-tuning.
For the rest of the cases, we used the same baseline which was trained using horizontal augmentation in the pixel space but the transfer learning parts are different. 
In the second case, no augmentation has been applied while in another experiment, we applied the augmentations in the pixel space before the feed-forward step. 
Finally, our proposed method, augmentation in the embedding space is applied. The original image is fed through $\Phi$.
For an arbitrary augmentation function $g_j$, the embeddings of the original image are transformed by corresponding $\Omega_j$.
This new embedding is then used to the fine-tune $\Psi$. The final accuracy of these three models after fine-tuning for 100 epochs is given in Table~\ref{tab:results} and the training curves are also demonstrated in Fig.~\ref{fig:transfer_learning_curves}. 
The results suggest that our method provides slightly lower accuracy than the augmentation in the pixel space baseline, but far better than fine-tuning without any augmentation.  

\begin{table}[t]
\caption{The accuracy of base and target models trained on CIFAR-100 and CIFAR-10 respectively. The augmentation setup is summarized as [Aug. for training of the base model]-[Aug. for fine-tuning the classifier ($\Psi$) during transfer learning]}
  \centering
\label{tab:results}
\begin{tabular}{|l|c|c|c|}
\hline
Network & VGG-16 & ResNet-18 & Inception-v3\\ \hline
\textbf{Augmentation} & \multicolumn{3}{c|}{\textbf{Base Network (CIFAR100)}} \\ \hline 
None                & 64.43\% & 61.89\% & 67.87\% \\ \hline
Hor. flip           & 71.47\% & 74.48\% & 78.20\% \\ \hline
Hor. \& Ver. flip   & 71.47\% & 72.46\% & 75.85\% \\ \hline
\textbf{Augmentation} & \multicolumn{3}{c|}{\textbf{Transfer Learning (CIFAR10)}} \\ \hline 
[Pixel]-[Pixel]     & 64.44\% & 78.87\% & 83.98\%  \\ \hline
[Pixel]-[None]      & 62.20\% & 76.25\% & 82.22\%  \\ \hline
[Pixel]-[Embed.]    & 63.68\% & 78.03\% & 82.20\%  \\ \hline
[None]-[None]       & 56.31\% & 65.46\% & 75.23\%  \\ \hline
\end{tabular}
\end{table}

The required computation for the FC based $\Omega$s and $\Psi$ sub-networks are negligible compared to the feed-forward pass of the transferred network. Therefore, if the network is required to be trained with the original image and only one other augmentation, the fine-tuning computation mainly consists of extracting the embeddings of the original data. 
As we are not extracting the embedding of the augmented input using our transferred network, almost $2\times$ saving in the computation can be achieved for each augmentation function.
\section{Conclusions and Future Work}\label{sec:conclusion}
In this paper, we presented a method for reducing the cost of data augmentation during the transfer learning of neural networks on embedded devices. 
The results show that our method reduces the computation drastically while the accuracy is negligibly affected. 
As future work, more complex augmentations and the effect of series of basic augmentations in the embedding space can be addressed.

\vspace{2mm}
\section*{Acknowledgement}
This work has been done during the internship of Mohammad Saeed Abrishami and Amir Erfan Eshratifar at Clarifai Inc. 
This research was also sponsored in part by a grant from the Software and Hardware Foundations (SHF) program of the National Science Foundation (NSF). 

\bibliographystyle{IEEEtran}
\bibliography{IEEEabrv,references}

\end{document}